# Coding historical causes of death data with Large Language Models

Bjørn Pedersen, Maisha Islam, Doris Tove Kristoffersen, Lars Ailo Bongo, Eilidh Garrett, Alice Reid, Hilde Sommerseth

**Abstract.** This paper investigates the feasibility of using pre-trained generative Large Language Models (LLMs) to automate the assignment of ICD-10 codes to historical causes of death. Due to the complex narratives often found in historical causes of death, this task has traditionally been manually performed by coding experts. We evaluate the ability of GPT-3.5, GPT-4, and Llama 2 LLMs to accurately assign ICD-10 codes on the HiCaD dataset that contains causes of death recorded in the civil death register entries of 19,361 individuals from Ipswich, Kilmarnock, and the Isle of Skye from the UK between 1861-1901. Our findings show that GPT-3.5, GPT-4, and Llama 2 assign the correct code for 69%, 83%, and 40% of causes, respectively. However, we achieve a maximum accuracy of 89% by standard machine learning techniques. All LLMs performed better for causes of death that contained terms still in use today, compared to archaic terms. Also they perform better for short causes (1-2 words) compared to longer causes. LLMs therefore do not currently perform well enough for historical ICD-10 code assignment tasks. We suggest further fine-tuning or alternative frameworks to achieve adequate performance.

## 1 Introduction

In historical demography, one of the main tasks is converting the existing knowledge bases of historical registers and microdata into encoded form, so that they can be easier utilised for research purposes. Many classification systems have been created over the years to help standardise the converted data, such as The Historical International Standard of Classification of Occupations (HISCO) [1] for occupational information, or the International Classification of Diseases (ICD) [2] for health information. However, a shared, defining trait of all of these classification systems is that, when applied to historical data, the encoding requires a tremendous effort in terms of manual work, usually done by domain experts.

      The aim of this paper is therefore to explore the possibility of using Large Language Models (LLMs) to automate the encoding of historical causes of death into the ICD-10 system.[1] In the WHO's current ICD-10 classification system, there are over 14,000 unique codes in the base version [4], but several countries have their own, extended versions of previous revisions of the system. Additionally, there are several codes with a high degree of overlap, making it

---

[1] Initially, the primary goal of the study was to assign ICD-10h codes, a version of ICD-10 created for use on historical causes. However, due to the unavailability of the ICD-10h system online, we concentrated on ICD-10. We believe the findings of this study are transferable to ICD-10h when it is accessible to LLMs. [3]

challenging to distinguish one disease or injury from another, for both human experts and automated systems.

LLMs like GPT (Generative Pre-trained Transformer) and Llama are artificial intelligence models built to both understand and generate human-like text. They are based on the transformer deep learning architecture presented by Vaswani et al. in 2017 [5]. They have multiple uses, in particular within the field of Natural Language Processing (NLP) [6–8], and have shown great potential for understanding text and for classification problems [9, 10]. *Prompt engineering* is a key feature for LLMs. All models require a prompt, which is the input to the model provided by a user through natural text, and by combining this with a set of instructions as to how the model should behave we can expect model performance to increase [11]. It is even possible to provide the models with highly specific examples of desired behaviour and outputs [12].

A notable drawback with LLMs is their propensity to "hallucinate" [13], meaning that the models present plausible outputs that are falsehoods. Also, LLMs are black boxes and therefore lack the needed transparency to explain their responses. Other machine learning (ML) models can provide a confidence metric for how confident the underlying algorithm is about a predicted value. This is not possible for an LLM because it does not predict an answer to the input text. Rather, it generates a response text by continually predicting which token should be next in the sentence it is constructing [14].

While LLMs have previously been used to generate ICD-10 billing codes[2] from modern hospital datasets, using them to classify historical causes of death remains a novel and worthwhile case study. We evaluate how current LLMs would perform for this specialised task of classifying a dataset containing a mixture of historical and currently used terms for a variety of causes of death.

## 1.1 Previous work

Up until now the work of assigning ICD codes to historical causes of death has been a process done manually by domain experts [15–18], but state-of-the-art ICD classifications for contemporary datasets [19–21] are usually achieved using pre-trained encoder transformers such as BERT [22] or automated coding systems such as ACME [23, 24]. Recently, we have seen an increased interest in exploring the capabilities of LLMs for ICD classification and other health data related tasks. Soroush et al. [25] have assessed the performance of OpenAI's GPT-3.5 and GPT-4 when generating ICD billing codes, but found that the LLM tendency to "hallucinate" key details would present too much of a problem for actual implementation in a healthcare context. Boyle et al. [26] created a novel tree search approach, guided by the LLM and based on the ICD code description, and managed to achieve competitive results without doing any task-specific training.

---

[2] Billing codes are standardised codes used in the healthcare industry of some countries to represent various medical procedures, treatments and services. These codes serve as a means of communication between healthcare providers, insurance companies and regulatory bodies, facilitating the billing and reimbursement process for medical services.

# 2 Methods

## 2.1 Historical Causes of Death (HiCaD) dataset

The dataset used in this project was created by a team at the Cambridge Group for the History of Population (CAMPOP) at the University of Cambridge. It covers three areas of the UK, the town of Ipswich in England, and the town of Kilmarnock and the Isle of Skye, both in Scotland. It spans the period 1861 to 1901, and includes 45,687 individual registered deaths. There are only infant deaths (i.e. a child who died prior to their first birthday) for Ipswich, while Skye and Kilmarnock also have deaths from other age groups. hence roughly ⅓ of the total deaths are those of infants. Historic causes of death are notoriously difficult to code, since there are a large number of illegible entries as well as archaic terms, vague causes and symptoms. Although it was a legal requirement for the cause of death to be certified by a doctor in both England and Scotland at this time, this did not universally happen, particularly when the deceased had not been treated by a doctor during their last illness. In such cases, a cause of death suggested by the informant (usually a relative) might have been recorded instead. Patterns of medical treatment mean that the deaths of the very young, the very old, and those who died from accidents or acute conditions were less likely to have been allocated a cause by a doctor. On the other hand, the ease of identifying particular causes, such as certain infectious diseases with very characteristic marks like the distinctive rash from smallpox or the red, swollen tongue of scarlet fever, means that the reporting of deaths from such causes may be more reliable, even if the informant was a lay person [27, 28].

This dataset was manually constructed, in batches, by two domain experts over at least a decade. It has been coded to both the 10th revision *of the International Statistical Classification of Diseases and Related Health Problems* (ICD-10) as well as a variant of the ICD-10 called the ICD-10h, to accommodate causes of death found in historical populations. The latter was originally developed for the Digitising Scotland project, based at the University of Edinburgh [29]. Currently it is being expanded for European comparisons by the SHiP+ network [3].

The ICD-10h version is currently only available offline and therefore not part of any LLM training set. This makes it not feasible to directly use current commercial LLMs to classify causes of death into the ICD-10h system without fine-tuning or embedding the knowledge into the model. However, since each cause of death in the dataset was given both a corresponding ICD-10 and ICD-10h code, we can instead use the ICD-10 code as the target for classification.

From this dataset of 45,687 registered deaths, we constructed a smaller dataset of cause of death strings by randomly sampling 19,361 individuals. This was done to reduce the cost of using LLMs. We will refer to this smaller dataset as the HiCaD (Historical Causes of Death). The original dataset consisted of 21 variables. In addition to personal information about the individual (sex, age at death, length of last illness), it also contained the original cause of death text string from the death registration. This string could contain multiple diseases suffered and/or injuries sustained by the person, which were thought to contribute to their death. Each one of these distinct diseases and/or injuries had been separated out, standardised, and given an ICD-10h code, by the original domain experts. For the HiCaD dataset however, we only kept the variables that were necessary for doing predictions through the LLMs and to analyse the results; the original cause of death text string, the first injury/illness reported as a cause of death and its corresponding ICD-10/ICD-10h code, and finally the historic category of disease (e.g. airborne disease, water- and foodborne disease).

## 2.2 The hierarchical structure of ICD-10 codes

The ICD-10 coding system represents specific diseases or injuries as an alphanumeric code, usually up to 4 characters long. The ICD classification system is constructed using a hierarchy with 5 levels (Table 1). The first level corresponds to a chapter within the classification system, but is not represented as a character in the final code. Levels 2 and 3 serve to narrow down the type or location, and are represented by the first 3 characters in the code, called blocks. Most of the three-character categories are subdivided by means of a fourth, numeric character after a decimal point, allowing up to 10 subcategories (0-9); this is the fourth level. The final level is the actual code that is given to the disease or injury.

**Table 1.** Overview of the hierarchical structure of the ICD-10 system, using the example code *A15.1*.

| Level | Level name | Code range | Description |
|---|---|---|---|
| 1 | Chapters | I-XXII | Each chapter represents a type of disease or injury |
| 2 | Blocks | A00-B99 | Certain infectious and parasitic diseases |
| 3 | Categories | A15-A19 | Tuberculosis |
| 4 | Subcategories | A15.0-A15.9 | Respiratory tuberculosis, bacteriologically and histologically confirmed |
| 5 | Code | A15.1 | Tuberculosis of lung, confirmed with culture only |

Within the ICD-10 and ICD-10h coding systems, there exist variables that categorise causes of death by type. For the ICD-10 system this is the chapter, and the ICD-10h equivalent is called *Historical category.* Late twentieth and twenty-first century data can be accurately classified using the ICD10 chapters, but chapters aren't always appropriate or useful for classifying historic causes of death. This is partly because codes were assigned to words or terms which could change meaning over time and partly because the lack of specificity in historic causes means that many historic causes end up in the 'Signs and Symptoms' Chapter of ICD10. Histcat offers a more historically sensitive classification which ensures codes are grouped in ways which reflect 19th century usage and knowledge.

## 2.3 LLMs

We employ OpenAI's default GPT-3.5-turbo and GPT-4 (Generative Pre-trained Transformer), as well as Meta's Llama 2 model, specifically the Llama-2-13b-chat-hf. We ran the experiments between November 1st and 8th in 2023. We kept the model hyperparameters at a default level in order to get the models' base level performance on the task, but we also employed prompt engineering to explore how this impacts the LLMs' performance.

## 2.4 Prompt engineering

The way a prompt is phrased has a massive impact on the LLM output. The models are also very susceptible to the tone of the user's feedback. If a user says that the answer is incorrect, the model changes its answer. Hence, we found that a set of very clear and concise instructions on how to behave, in a very neutral tone, works best. We included the prompt as part of the input for each request to classify a cause of death, as we found that the model would start disregarding the specific instructions of the prompt over time if we did not. The prompt[3] we used is:

{'role' : 'system', 'content' : """"Assistant is an intelligent chatbot designed to help the user assign clinical ICD-10 codes to causes of death.
    Instructions:
    - Only answer using standard ICD-10 codes, do not use ICD-10-CM billing codes.
    - Only return a single ICD-10 code per injury and/or disease found in the given cause of death.
    - Each ICD-10 code should not consist of more than 5 characters, the typical format looks like this: 'X01.0'
    - Your answers should be in the following format: 'Cause of death: <CAUSE OF DEATH>, ICD-10 code: <ICD-10 CODE>'
    - If you are unsure of an answer, do not try to guess. Instead, write the following reply: 'Cause of death: Unknown, ICD-10 code: Æ99.9'.
    """}

## 2.5 Experiments

**Correct classification.** In the first experiment, we measured how many correct classifications the models achieved on the dataset. We used two ways to define a correct classification. The first is called a *full match*, and is achieved when the model output corresponds exactly with the ICD-10 code manually assigned by the domain experts. The second type of correct classification is called a *partial match*, and is defined as having the first 3 characters of the model's output match the first 3 characters of the ICD-10 code assigned by the domain expert. We count this as correct as each ICD-10 code only has a maximum of 10 possible subcategories (0-9) which make up the 4th character in the code, so the required manual work to find this last character will be relatively easy.

    As part of the experiment, we count the number of times where a LLM replies that it could not classify the cause of death, as opposed to hallucinating a wrong code. To do this, we defined a specific error code as part of the prompt we gave to each model, 'Æ99.9', with instructions to return this code if it could not confidently classify the cause of death.

---

[3] The prompt shown here follows the format required by OpenAI's GPT models. Other types of LLMs may have different structures and syntax.

**Comparison with simpler methods.** The second experiment aimed to compare the results of the LLMs with two alternative classification techniques. The first alternative was the traditional Machine Learning-based models, Random Forest and Support Vector Machine (SVM). Random Forest is an ensemble learning method that combines multiple decision trees to enhance predictive accuracy. It minimises overfitting and captures complicated relationships well as it builds each tree using bootstrapped data and takes random feature subsets into account. SVM is a robust algorithm for classification and regression. It seeks the optimal hyperplane to maximise the margin between classes in feature space. SVM handles both linear and non-linear relationships, offering effective solutions in high-dimensional spaces with clear class boundaries. We incorporated a grid search to find the best set of hyperparameters for each model.

The second alternative was a basic string similarity comparison method, comparing the HiCaD dataset with a dictionary of standardised causes of death, originally created by domain experts.

**Temporal context of the causes of death.** We expected that LLMs would perform worse on terms that only exist in historical registers or have a different meaning from the current understanding of the term. Therefore, our third experiment aimed to compare the number of correct classifications done by the LLMs for both types of terms. We sorted our causes of death into what we refer to as "archaic" and "current" causes, using the manually coded ICD-10h code. If the final character in the ICD-10h code is a 0, it means that this is a cause of death term that exists in both contemporary and historic registers. If the ICD-10h code ends in any other digit, then the cause of death term is only found in historical registers, or the term might be understood differently today.

**Complexity of the input values.** Over 80% of the original cause of death text strings within the HiCaD dataset consist of 3 words or less, but the remaining 20% cover a range of 4-41 words. In our fourth experiment, we calculated if the number of words in the input text had any impact on the models' ability to correctly assign the cause of death to an ICD-10 code.

To do this, we grouped the causes of death by word count as *short* (1-2 words), *medium* (3-4 words), and *long* (5+ words) and measured the error rate for each group. These cutoffs were chosen as causes with one or two words are likely to represent just a single cause of death (e.g. 'pneumonia', 'scarlet fever') but causes with more words are more likely to represent accidents and multiple causes of death (e.g. 'pneumonia following measles').

To measure the agreement between the three LLM models, that is, the extent to which all LLMs assigned the same code for each cause of death, the Fleiss Kappa [30] was calculated. This was done per word count category and for overall match/no match agreement between the models. Values above 0.2 indicate fair agreement, above 0.4 moderate, above 0.6 substantial, above 0.8 almost perfect.

**Histcat classification.** Each ICD-10 code can only belong to a single historical category (Histcat). In the fifth experiment we calculated if the distribution of ICD-10 codes within the

Histcats in the manually coded HiCaD dataset matched the distribution in the models' outputs. This tells us if there are any particular types of Histcats that the LLMs are worse at classifying than others.

# 3 Results

## 3.1 Correct classification

These results are summarised in Table 2. For the HiCaD dataset, consisting of 19,353 cause of death text strings, GPT-3.5 coded the causes as a full match in 31% and a partial match in 38%, with no overlap, meaning that 69% of all causes were encoded correctly, and 31% of the causes were given an incorrect ICD-10 code. GPT-4 achieved 58% for full matches and 25% for partial matches, leading to 83% of all causes of death being encoded into the correct ICD-10 code, with an error rate of 17%. Llama 2 coded 9% of all causes as a full match, 31% as a partial match, and had an error rate of 60%. These results show that GPT-4 outperforms GPT-3.5 in the accurate classification of ICD-10 codes, and both models show significantly better performance than Llama 2.

GPT-3.5 had 1,101 causes of death that received the error code we defined as part of the model prompt, which corresponds to 19% of all errors made by GPT-3.5. For GPT-4, this happened for 247 causes of death, or 1% of all GPT-4 errors. For Llama 2, none of the errors were assigned the specified error code.This means that GPT-4 and Llama 2 are more likely to predict a wrong ICD-10 code than to answer with the error code, whereas GPT-3.5 showed the opposite. GPT-4's errors are broken down into 16% hallucinations and 1% error code, but GPT-3.5's errors were 12% hallucinations and 19% error code, and for Llama 2 all errors were hallucinations.

Additionally, some model outputs were incorrectly formatted, ignoring the specifications of the prompt instructions and as a result it became impossible to extract an ICD-10 for the cause of death. For GPT-3.5 this happened in 908 cases, or 4.7% of all errors. For GPT-4, 535 cases or 2.7% of all errors. For Llama 2, 77 cases or 0.4% of all errors.

**Table 2.** An overview of how LLMs encoded the historical causes of death.

| Model | Full match | Partial match | Correct (full + partial) | Errors |
| --- | --- | --- | --- | --- |
| GPT-3.5 | 31% | 38% | 69% | 31% |
| GPT-4 | 58% | 25% | 83% | 17% |
| Llama 2 | 9% | 31% | 40% | 60% |

## 3.2 Comparison with alternate solutions

Random Forest achieved an accuracy of 87% and SVM achieved an accuracy of 89% on a test set of 19,360 cause of death text strings, sampled from the full dataset of over 45,000 rows.

When using string similarity comparison on the HiCaD dataset, consisting of 19,353 rows, the number of correctly coded causes of death were 6%, meaning that 94% of all causes were assigned an incorrect ICD-10 code.

These results show that both Random Forrest and SVM outperforms the LLMs, with the best result obtained by an LLM being 83% correct classification. It does, however, also show that LLMs vastly outperforms string similarity comparison.

## 3.3 Temporal context of the causes of death

Within the HiCaD dataset, we found that 9,773 rows consisted of causes of death that could be categorised as archaic, and 9,580 causes as current. This gives a split of 50.5% to 49.5%. With defining both a full and partial match as correct, we found that GPT-3.5 achieved a score of 55% correct for archaic causes of death, and 83% for current causes. GPT-4 classified 75% of cases correctly for archaic causes and 90% correct for current. Llama 2 managed to correctly code 35% of the archaic causes, and 45% of current causes.

## 3.4 Complexity of input values

The error rate of the LLM models per word category 1-2, 3-4 and 5+ was 26%, 39% and 38% for GPT-3.5, 15%, 18% and 29% for GPT-4, and 55%, 68% and 66% for Llama 2, respectively. Showing that GPT-4 performed better than GPT-3.5 and both outperform Llama 2, this holds for all categories.

When comparing the degree of agreement between the LLMs, i.e. to which extent all models gave the same output, Fleiss Kappa was lowest for the medium (3-4 words) category with 0.19 meaning slight agreement, for short (1-2) words it was 0.27 which indicates fair agreement. Agreement was fair for the long (5+ words) category with 0.31, meaning that the models tended to most often give the same output for the causes of death that consisted of 5+ words, and the least often for the causes of death that were 1-2 words long. Overall, the models' agreement was measured as 0.27, indicating a fair level of agreement.

## 3.5 Histcat classification

Figure 1 shows the causes of death which were classified incorrectly by GPT-4, see Table 2, and how these are distributed along the Histcat classification scheme. The diagonal row displays the cases where even though the model assigned an incorrect ICD-10 code to the cause of death, that wrong code still falls in under the correct Histcat.
For example, in the Histcat 'Childbirth', 64% of causes were assigned a full or partial match by the LLM, which is displayed within the parentheses following the Histcat name on the y-axis. Of the remaining 36% of causes which were classified incorrectly, 54% were classified into the correct Histcat but an incorrect ICD-10 code, while 17% were classified into the Histcat 'Perinatal', 8% were placed within "Ill defined", etc.

GPT-4 showed strong performance across most categories, except for 'Violence', where it only classified the cause to the correct ICD-10 code in 37% of cases. Despite this, it managed to classify those causes into the correct Histcat 89% of the time. The opposite can be seen in the 'Debility' category, where GPT-4 managed to find the correct ICD-10 code in 59% of cases, but 0% of the errors for these codes ended up in the correct Histcat. Figures for GPT-3.5 and Llama 2 are in the appendix.

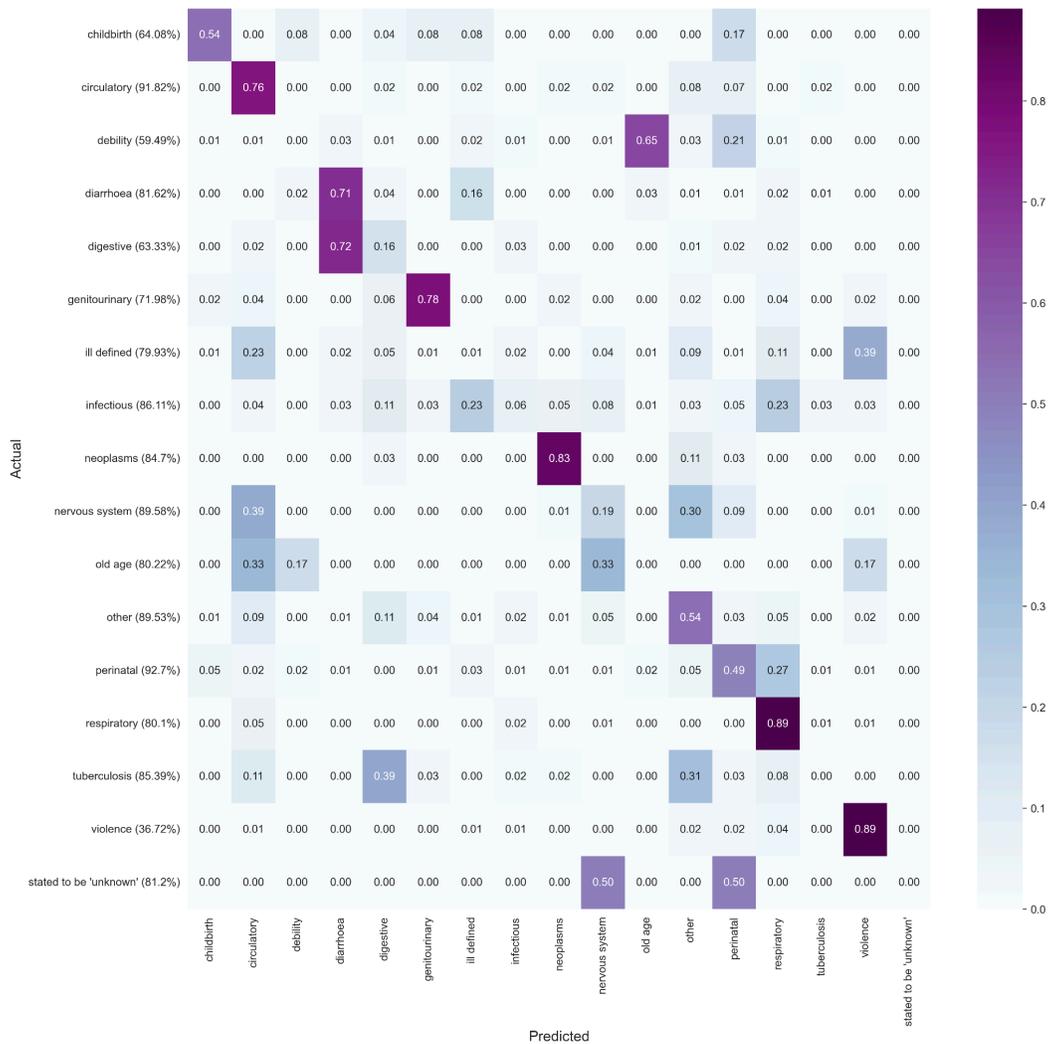

**Figure 1:** A heatmap of wrong classifications made by GPT-4 in regards to Histcat. If the model classified the cause of death into the wrong ICD-10 code, but the erroneous code still belongs to the same category, then the errors will cluster on the diagonal. If not, the errors will spread out along the row.

# 4 Discussion

## 4.1 Main findings

We explored how well "out of the box" Large Language Models performed on classifying ICD-10 codes for historical death records from three different areas of the UK. GPT-4 performed the best in terms of coding accuracy among GDPR-compliant LLMs, notably surpassing GPT-3.5 and Llama 2, especially for causes of death using current terms. All models showed improved performance with shorter causes of death (1-2 words). While model agreement varied with cause length, showing average agreement for longer and poor for medium-length causes, overall substantial agreement was observed.

Common errors in model classifications were linked to three patterns: abbreviations in the source data (e.g., "con" for "consumption"), coding into the chapter of 'Symptoms, signs and abnormal clinical and laboratory findings, not elsewhere classified' which is known to be more difficult to code, and specific terms like 'teething' and 'dentition' that are less common in modern datasets but prevalent in historical infant mortality records.

In the Histcat classification, the uneven performance we observed may relate to the varying number of codes per Histcat, as categories with more codes like 'Violence' (3,278 unique ICD-10 codes) have a higher likelihood of at least partial classification success than a category with less codes like 'Debility' (3 unique ICD-10 codes).

All LLMs were outperformed by traditional machine learning models. Despite this, the risk of overfitting [31] remains a concern due to the dataset's imbalance and numerous labels. Nevertheless, all models did surpass the basic string similarity approach using Jaro-Winkler distance.

## 4.2 Comparison to related work

Our findings mostly correlate with those of Soroush et al. [25], in that we have found the performance of base level LLM models to be inadequate for practical use when it comes to assigning ICD codes to historical causes of death. In our results, the majority of errors did not contain the error code we defined as part of our model prompt, which we instructed the models to use if it did not "know" the answer. Instead, the majority of the errors would fall in under the definition of hallucinations, i.e. creating plausible but incorrect statements [13].

We introduced the concept of full matches and partial matches, and we saw that all models had a potential for achieving more correct classifications if they had managed to find the correct subcategory for the cause of death. This could have potentially been done in a similar manner to what Boyle et al. [26] term "meta-refinement of predicted codes". They used a tree-search approach to find all possible, relevant codes for their data, and then asked GPT-4 to remove false positives. In Section 4.3, we discuss the possibility of prompting the LLMs a second time, to find the correct subcategory based on the first 3 characters of the ICD-10 code that were given as output during the first round of prompting.

## 4.3 Future work

For future projects where the goal will be to achieve the best coding rate possible using LLMs on historical causes of death, one technique to use is *fine-tuning*, a type of transfer learning [31], where pre-trained models are further trained on new, labelled data [32]. This would entail downloading a pre-trained LLM and then fine-tune it on a dataset containing causes of death. Based on previous studies, we expect that this will yield improved results [33, 34].

As seen in Section 3, all models gave partial matches as part of their outputs, indicating that if the last character had also been correct, these causes would have been a full match. One possible solution, that can be built upon our current work, is to automate a solution where each cause of death is passed in to a LLM for a second round of classification, this time presenting the model with both the cause of death text string, the almost completed code, as well as a list of possible subgroups that the model is then asked to choose between; this is an approach similar to Boyle et al.'s "meta-refinement of predicted codes".

Another promising solution is Retrieval-Augmented Generation (RAG) [35, 36]. RAG is an AI framework that can improve LLM responses by grounding the model on external knowledge bases that the users supply. Using RAG, we would be able to use the ICD-10h master list of terms and codes as a knowledge base, and when prompting an LLM to classify a cause of death into an ICD-10h code, it would first retrieve information from the knowledge base before responding. This would increase the probability of getting a correct response and it would give the users a source for the answer.

## 4.4 Limitations

We intended to use LLMs for assigning ICD-10h codes to the causes of death, as it is tailored for historical terms, but resorted to standard ICD-10 codes due to the ICD-10h's unavailability online, potentially affecting the number of full matches. However, ICD-10h codes have an additional two digits which might have also increased error risks. Furthermore, the LLMs' tendency to align with the ICD-10-CM billing system, despite instructions to avoid it, contributed to the low full match count.

We provided models with original cause of death text strings, which often includes several diseases and/or injuries that the individual was suffering from at the time of death, instructing them to return an ICD-10 code for each. Our analysis compared the first code from the models to the first in our manual dataset, as different causes are not often easy to distinguish. The LLMs may have focused on the true underlying cause mentioned later in the text, as prompted, rather than the first listed condition.

Another limitation is that the analysis was done on records of individual deaths, not on unique strings. This has the advantage of assessing how many deaths were likely to be correctly coded, but not necessarily how well particular commonly written strings were coded. Additionally, the dataset was skewed towards infant deaths.

At the time of the experiment, a more powerful version of the Llama 2 model was trained on 70 billion parameters. However, we could not use this version of the model, as we did not have a powerful enough computer to run it. Hence we were restricted to the 13 billion parameter model. This has most likely impacted the performance of Llama 2 reported on in this paper.

# 5 Conclusion

In this paper, we explored the use of Large Language Models (LLMs) to encode historical causes of death data. We experimented with the current default versions (The experiments were run in early November 2023) of OpenAI's GPT-3.5 and GPT-4 models, as well as Meta's Llama 2, on a dataset covering the period 1861-1901 gathered from Ipswich, Kilmarnock, and the Isle of Skye in the UK. We found that no LLM performed this task satisfactorily, with GPT-4 achieving a correctness-score of 82.6% at the highest, but it was only able to correctly match the manual coding completely in 57.9% of cases. We found that all LLMs achieved better results for causes of death where more current terms are used, than for archaic terms. They also performed better for causes of death that were composed of shorter text strings. We found that the models made more errors when given longer text strings, but that these errors had a high degree of agreement between models, meaning that the models all tended to make the same error.

We compared the performance of the LLMs to both classical machine learning methods such as a random forest classifier and SVM, and a simple string similarity matching algorithm. We found that while the LLMs performed much better than string similarity matching, they still perform worse than the machine learning methods.

It is important to keep in mind that these results reflect the use of LLMs that were trained to be chatbots with extensive general knowledge, rather than models that have been specifically pre-trained or fine-tuned with domain expertise in historical causes of death or the ICD-10 system. As such, we have chosen to treat these results as a baseline that we can use to compare with for future projects where we utilise fine-tuning or frameworks like Retrieval-Augmented Generation..

**Acknowledgement.** We wish to thank the General Register Office for Scotland (GROS), now National Records of Scotland (NRS), for special permission to transcribe the contents of the civil registers of the Isle of Skye and the town of Kilmarnock from 1855-1901, which was carried out under the ESRC funded project *Determining the Demography of Victorian Scotland Through Record Linkage* (RES-000-23-0128) held at the Cambridge Group for the History of Population and Social Structure, University of Cambridge. We acknowledge funding provided by the Research Council of Norway through the infrastructure project Historical Registers (project number 322231).

**Code and data availability.** The dataset used in this study is not publicly available, but all our code and the LLM prompts are available and open source using the MIT licence. It can be found at
https://github.com/HistLab/Coding-historical-causes-of-death-data-with-Large-Language-Models

# References

1. Van Leeuwen, M.H.D., Maas, I., Miles, A.: HISCO: Historical International Standard Classification of Occupations. Leuven University Press, 2002 (2002).
2. Hirsch, J.A., Nicola, G., McGinty, G., Liu, R.W., Barr, R.M., Chittle, M.D., Manchikanti, L.: ICD-10: History and Context. Am. J. Neuroradiol. 37, 596–599 (2016). https://doi.org/10.3174/ajnr.A4696.
3. Janssens, A.: Constructing SHiP and an International Historical Coding System for Causes of Death. Hist. Life Course Stud. 10, 64–70 (2021). https://doi.org/10.51964/hlcs9569.
4. WHO | FAQ on ICD, https://web.archive.org/web/20041017011702/http://www.who.int/classifications/help/icdfaq/en/, last accessed 2023/11/06.
5. Vaswani, A., Shazeer, N., Parmar, N., Uszkoreit, J., Jones, L., Gomez, A.N., Kaiser, Ł., Polosukhin, I.: Attention is All you Need. In: Guyon, I., Luxburg, U.V., Bengio, S., Wallach, H., Fergus, R., Vishwanathan, S., and Garnett, R. (eds.) Advances in Neural Information Processing Systems. Curran Associates, Inc. (2017).
6. Wang, L., Lyu, C., Ji, T., Zhang, Z., Yu, D., Shi, S., Tu, Z.: Document-Level Machine Translation with Large Language Models, http://arxiv.org/abs/2304.02210, (2023). https://doi.org/10.48550/arXiv.2304.02210.
7. Wang, S., Sun, X., Li, X., Ouyang, R., Wu, F., Zhang, T., Li, J., Wang, G.: GPT-NER: Named Entity Recognition via Large Language Models, http://arxiv.org/abs/2304.10428, (2023). https://doi.org/10.48550/arXiv.2304.10428.
8. Zhang, B., Yang, H., Zhou, T., Babar, A., Liu, X.-Y.: Enhancing Financial Sentiment Analysis via Retrieval Augmented Large Language Models, http://arxiv.org/abs/2310.04027, (2023). https://doi.org/10.48550/arXiv.2310.04027.
9. Sun, X., Li, X., Li, J., Wu, F., Guo, S., Zhang, T., Wang, G.: Text Classification via Large Language Models, http://arxiv.org/abs/2305.08377, (2023). https://doi.org/10.48550/arXiv.2305.08377.
10. Chen, S., Li, Y., Lu, S., Van, H., Aerts, H.J., Savova, G.K., Bitterman, D.S.: Evaluation of ChatGPT Family of Models for Biomedical Reasoning and Classification, http://arxiv.org/abs/2304.02496, (2023). https://doi.org/10.48550/arXiv.2304.02496.
11. Ahmed, T., Devanbu, P.: Few-shot training LLMs for project-specific code-summarization. In: Proceedings of the 37th IEEE/ACM International Conference on Automated Software Engineering. pp. 1–5. Association for Computing Machinery, New York, NY, USA (2023). https://doi.org/10.1145/3551349.3559555.
12. Brown, T.B., Mann, B., Ryder, N., Subbiah, M., Kaplan, J., Dhariwal, P., Neelakantan, A., Shyam, P., Sastry, G., Askell, A., Agarwal, S., Herbert-Voss, A., Krueger, G., Henighan, T., Child, R., Ramesh, A., Ziegler, D.M., Wu, J., Winter, C., Hesse, C., Chen, M., Sigler, E., Litwin, M., Gray, S., Chess, B., Clark, J., Berner, C., McCandlish, S., Radford, A., Sutskever, I., Amodei, D.: Language Models are Few-Shot Learners, http://arxiv.org/abs/2005.14165, (2020). https://doi.org/10.48550/arXiv.2005.14165.
13. Hallucination Definition & Meaning - Merriam-Webster, https://www.merriam-webster.com/dictionary/hallucination, last accessed 2023/11/20.
14. Bowman, S.R.: Eight Things to Know about Large Language Models, http://arxiv.org/abs/2304.00612, (2023). https://doi.org/10.48550/arXiv.2304.00612.
15. Bailey, M.J., Leonard, S.H., Price, J., Roberts, E., Spector, L., Zhang, M.: Breathing new life into death certificates: Extracting handwritten cause of death in the LIFE-M project. Explor. Econ. Hist. 87, 101474 (2023). https://doi.org/10.1016/j.eeh.2022.101474.
16. Revuelta-Eugercios, B., Castenbrandt, H., Løkke, A.: Older rationales and other challenges in handling causes of death in historical individual-level databases: the case of

Copenhagen, 1880–1881. Soc. Hist. Med. 35, 1116–1139 (2022). https://doi.org/10.1093/shm/hkab037.
17. Anderton, D.L., Leonard, S.H.: Grammars of Death: An Analysis of Nineteenth-Century Literal Causes of Death from the Age of Miasmas to Germ Theory. Soc. Sci. Hist. 28, 111–143 (2004). https://doi.org/10.1017/S0145553200012761.
18. Williams, N.: The Reporting and Classification of Causes of Death in Mid-Nineteenth-Century England. Hist. Methods J. Quant. Interdiscip. Hist. 29, 58–71 (1996). https://doi.org/10.1080/01615440.1996.10112730.
19. Chen, P.-F., Chen, K.-C., Liao, W.-C., Lai, F., He, T.-L., Lin, S.-C., Chen, W.-J., Yang, C.-Y., Lin, Y.-C., Tsai, I.-C., Chiu, C.-H., Chang, S.-C., Hung, F.-M.: Automatic International Classification of Diseases Coding System: Deep Contextualized Language Model With Rule-Based Approaches. JMIR Med. Inform. 10, e37557 (2022). https://doi.org/10.2196/37557.
20. Edin, J., Junge, A., Havtorn, J.D., Borgholt, L., Maistro, M., Ruotsalo, T., Maaløe, L.: Automated Medical Coding on MIMIC-III and MIMIC-IV: A Critical Review and Replicability Study. In: Proceedings of the 46th International ACM SIGIR Conference on Research and Development in Information Retrieval. pp. 2572–2582 (2023). https://doi.org/10.1145/3539618.3591918.
21. Huang, C.-W., Tsai, S.-C., Chen, Y.-N.: PLM-ICD: Automatic ICD Coding with Pretrained Language Models. In: Naumann, T., Bethard, S., Roberts, K., and Rumshisky, A. (eds.) Proceedings of the 4th Clinical Natural Language Processing Workshop. pp. 10–20. Association for Computational Linguistics, Seattle, WA (2022). https://doi.org/10.18653/v1/2022.clinicalnlp-1.2.
22. Devlin, J., Chang, M.-W., Lee, K., Toutanova, K.: BERT: Pre-training of Deep Bidirectional Transformers for Language Understanding. In: Burstein, J., Doran, C., and Solorio, T. (eds.) Proceedings of the 2019 Conference of the North American Chapter of the Association for Computational Linguistics: Human Language Technologies, Volume 1 (Long and Short Papers). pp. 4171–4186. Association for Computational Linguistics, Minneapolis, Minnesota (2019). https://doi.org/10.18653/v1/N19-1423.
23. Slik kodes og kvalitetssikres dødsårsaker i Dødsårsaksregisteret, https://www.fhi.no/op/dodsarsaksregisteret/dodsarsaken-kodes-med-icd-koder/, last accessed 2023/11/09.
24. Hernes, E., Johansson, L.A., Fosså, S.D., Pedersen, A.G., Glattre, E.: High prostate cancer mortality in Norway evaluated by automated classification of medical entities. Eur. J. Cancer Prev. 17, 331–335 (2008).
25. Soroush, A., Glicksberg, B.S., Zimlichman, E., Barash, Y., Freeman, R., Charney, A.W., Nadkarni, G.N., Klang, E.: Assessing GPT-3.5 and GPT-4 in Generating International Classification of Diseases Billing Codes, https://www.medrxiv.org/content/10.1101/2023.07.07.23292391v2, (2023). https://doi.org/10.1101/2023.07.07.23292391.
26. Boyle, J.S., Kascenas, A., Lok, P., Liakata, M., O'Neil, A.Q.: Automated clinical coding using off-the-shelf large language models, http://arxiv.org/abs/2310.06552, (2023).
27. Reid, A., Garrett, E.: Doctors and the causes of neonatal death in Scotland in the second half of the nineteenth century. Ann. Demogr. Hist. 2012, 149–179 (2013). https://doi.org/10.3917/adh.123.0149.
28. ALTER, G.C., CARMICHAEL, A.G.: Classifying the Dead: Toward a History of the Registration of Causes of Death. J. Hist. Med. Allied Sci. 54, 114–132 (1999). https://doi.org/10.1093/jhmas/54.2.114.
29. Scottish Historic Population Platform (SHiPP) | SCADR, https://www.scadr.ac.uk/our-research/shipp, last accessed 2023/11/09.
30. Fleiss, J.L.: Measuring nominal scale agreement among many raters. Psychol. Bull. 76,

378–382 (1971). https://doi.org/10.1037/h0031619.
31. Spring Research Presentation | College of Physical and Mathematical Sciences, https://web.archive.org/web/20070801120743/http://cpms.byu.edu/springresearch/abstract-entry?id=861, last accessed 2023/11/14.
32. 14.2. Fine-Tuning — Dive into Deep Learning 1.0.3 documentation, https://d2l.ai/chapter_computer-vision/fine-tuning.html, last accessed 2023/11/14.
33. Liga, D., Robaldo, L.: Fine-tuning GPT-3 for legal rule classification. Comput. Law Secur. Rev. 51, 105864 (2023). https://doi.org/10.1016/j.clsr.2023.105864.
34. Peng, B., Li, C., He, P., Galley, M., Gao, J.: Instruction Tuning with GPT-4, http://arxiv.org/abs/2304.03277, (2023). https://doi.org/10.48550/arXiv.2304.03277.
35. Lewis, P., Perez, E., Piktus, A., Petroni, F., Karpukhin, V., Goyal, N., Küttler, H., Lewis, M., Yih, W., Rocktäschel, T., Riedel, S., Kiela, D.: Retrieval-Augmented Generation for Knowledge-Intensive NLP Tasks, http://arxiv.org/abs/2005.11401, (2021). https://doi.org/10.48550/arXiv.2005.11401.
36. What is retrieval-augmented generation?, https://research.ibm.com/blog/retrieval-augmented-generation-RAG, last accessed 2023/11/16.

# Appendix

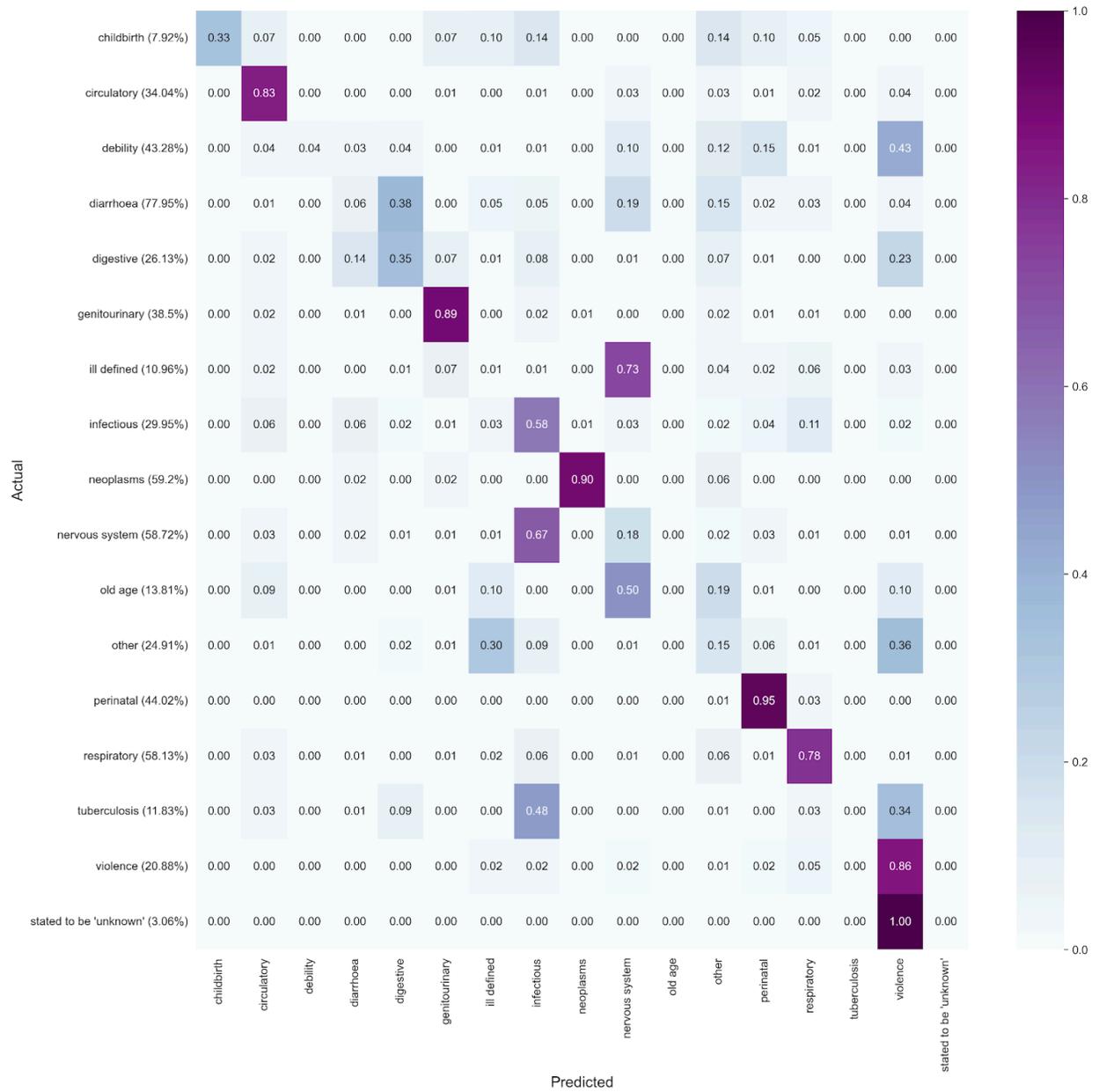

**Figure A1**. Llama 2 Histcat heatmap

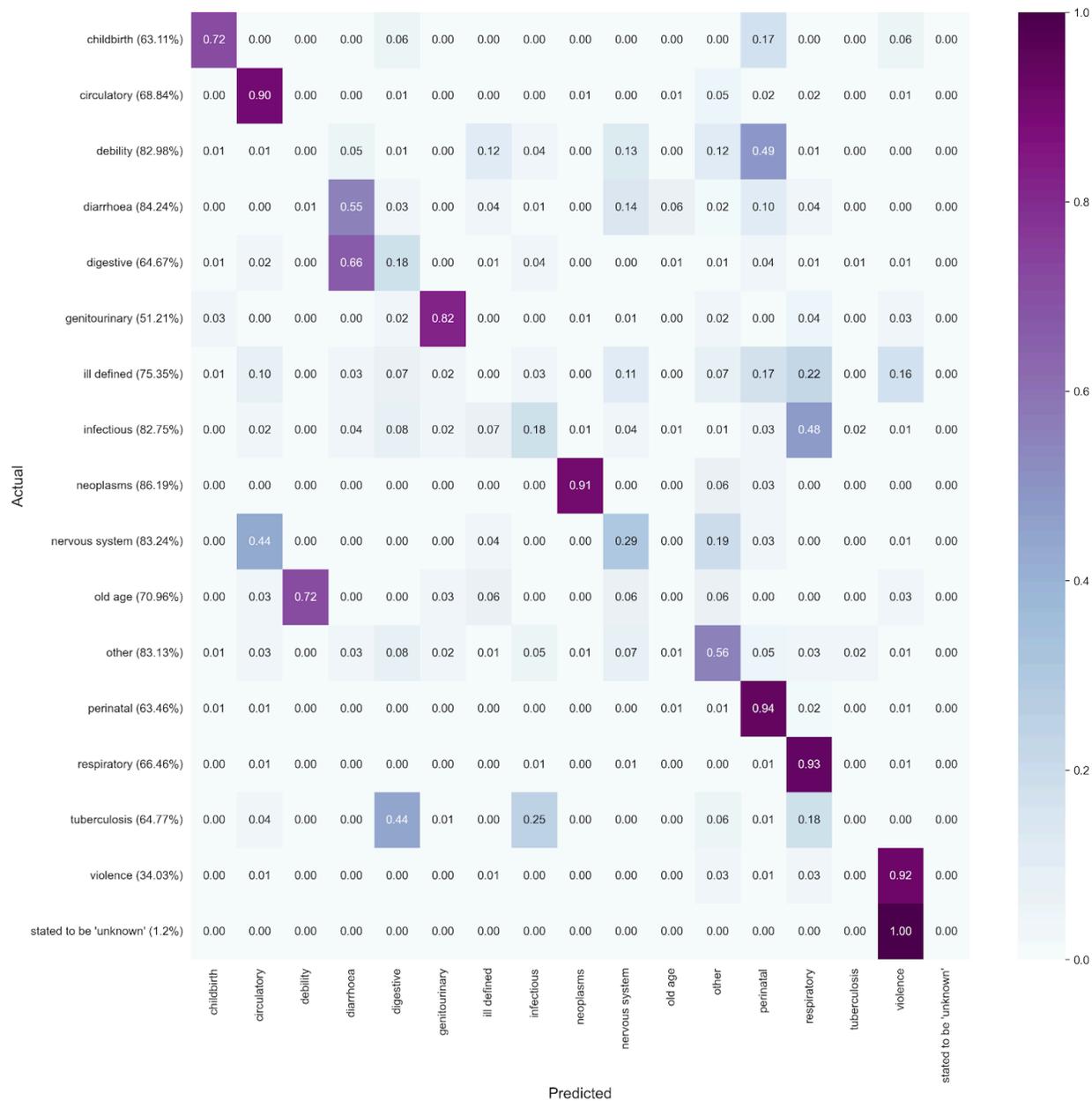

**Figure A2**. GPT-3.5 Histcat heatmap

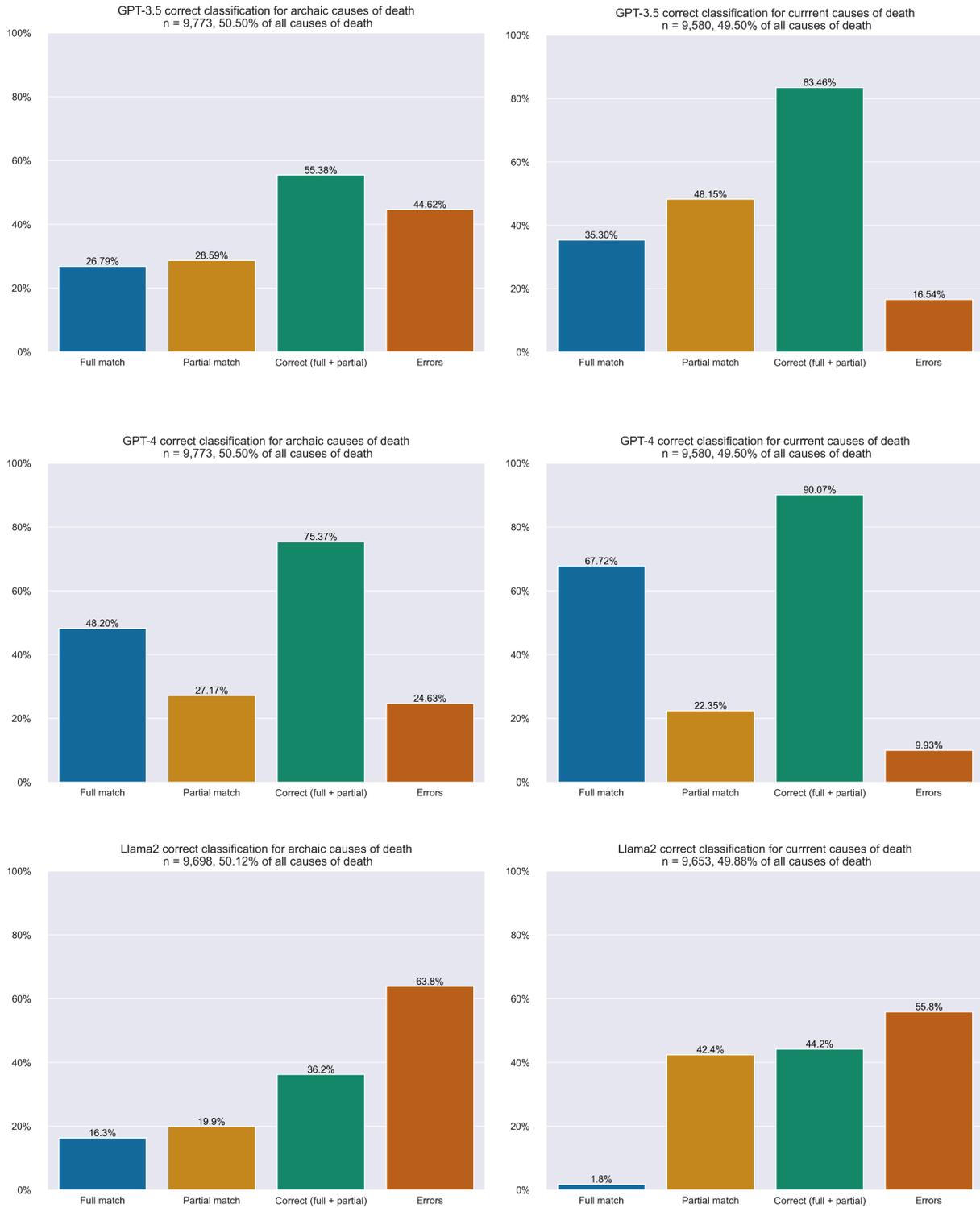

**Figure A3**. Classification results for archaic (left) and current (right) causes of death from GPT-3.5 (top row), GPT-4 (middle row) and Llama 2 (bottom row)